\documentclass{article} % For LaTeX2e
\PassOptionsToPackage{dvipsnames}{xcolor}
\usepackage{iclr2027_conference}
\usepackage[utf8]{inputenc}
\usepackage[T1]{fontenc}
\usepackage{times}
\usepackage{newunicodechar}
\newunicodechar{’}{\textquoteright}
\newunicodechar{‑}{-}

\usepackage{xcolor}
\usepackage{graphicx}
\usepackage{amsmath}
\usepackage{amssymb}   % for \checkmark
\usepackage{textcomp}  % for \textuparrow / \textdownarrow
\usepackage{booktabs}
\usepackage{longtable}
\usepackage{array}
\usepackage{adjustbox}
\newcolumntype{L}[1]{>{\raggedright\arraybackslash}p{#1}}

\usepackage{hyperref}
\usepackage{url}
\usepackage[capitalize,noabbrev]{cleveref}

\usepackage{xspace}
\makeatletter
\DeclareRobustCommand\onedot{\futurelet\@let@token\@onedot}
\def\@onedot{\ifx\@let@token.\else.\null\fi\xspace}
\def\eg{\emph{e.g}\onedot}

\def\ie{\emph{i.e}\onedot}

\makeatother

\title{SeeNav-Agent: Enhancing Vision-Language Navigation with Visual Prompt and Step-Level Policy Optimization}

\author{%
\begin{tabular}{@{}l@{}}
Zhengcheng Wang$^{*}$ \quad Zichuan Lin$^{*}$ \quad Yijun Yang \quad Haobo Fu \quad Deheng Ye \\[6pt]
\normalfont\fontsize{10.5}{14}\selectfont
Tencent AI Lab \\[3pt]
\normalfont\fontsize{10.5}{14}\selectfont
Email: \texttt{jensencwang@tencent.com}
\end{tabular}%
}

\iclrfinalcopy % Camera-ready: show authors and drop the review ruler.
\begin{document}

\maketitle
\lhead{Under review}
\begingroup
\renewcommand\thefootnote{*}
\footnotetext{Equal contribution.}
\endgroup

% ==========================================================================
% Section: Abstract  (from sec/0_abstract.tex of the ECCV source, verbatim body text)
% ==========================================================================
\begin{abstract}

Existing Vision-Language Navigation (VLN) agents based on Large Vision-Language Models (LVLMs) often suffer from errors of perception, reasoning, and planning, which significantly hinder their navigation performance. To address these limitations, a novel VLN agent framework, named SeeNav-Agent, is proposed. First, to reduce hallucinations of the visual module of the VLN agent, a dual-view Visual Prompt (VP) technique is introduced, which can also improve the agent’s understanding of current spatial states. Subsequently, a novel step-level Reinforcement Fine-Tuning (RFT) method, Step Reward Group Policy Optimization (SRGPO), is designed for post-training. In SRGPO, we first define verifiable process rewards for navigation, and then perform efficient step-level advantage estimation by randomly grouping navigation steps. SRGPO provides dense reward signals for the RFT of the VLN agent and enhances its planning capability. Experiments on both the EmbodiedBench Navigation (Embench-Nav) and VLN-CE validate the effectiveness of the proposed method. By introducing the zero-shot VP module, the GPT-4.1 achieves a navigation success rate of 86.7\%, surpassing the current best LVLM by approximately 20 percentage points (pp) on Embench-Nav and achieves an 11.1 pp improvement on VLN-CE compared to the base model. With VP+SRGPO, the Qwen2.5-VL-3B model reaches a navigation success rate of 72.3\%, outperforming the best existing LVLM model by 5.6 pp on Embench-Nav and achieves 8.9 pp improvement on VLN-CE compared to the baseline method. Moreover, compared to GRPO and GiGPO, the proposed SRGPO demonstrates significant improvements in training stability, convergence efficiency, and generalization capability.

\par\vspace{5pt}
\noindent\textbf{Code}\enspace
\href{https://github.com/WzcTHU/SeeNav-Agent}{github.com/WzcTHU/SeeNav-Agent}\par
\noindent\textbf{Model}\enspace
\href{https://huggingface.co/wangzc9865/SeeNav-Agent}{huggingface.co/wangzc9865/SeeNav-Agent}\par
\end{abstract}

% ==========================================================================
% Section: Introduction  (from sec/1_intro.tex of the ECCV source, verbatim)
% Only figure widths / paths were adjusted to fit AAAI's two-column letter layout.
% ==========================================================================
\section{Introduction}
\label{sec:intro}

\begin{figure}[t]
  \centering
%  \fbox{\rule{0pt}{2in} \rule{0.9\linewidth}{0pt}}
   \includegraphics[width=0.75\columnwidth]{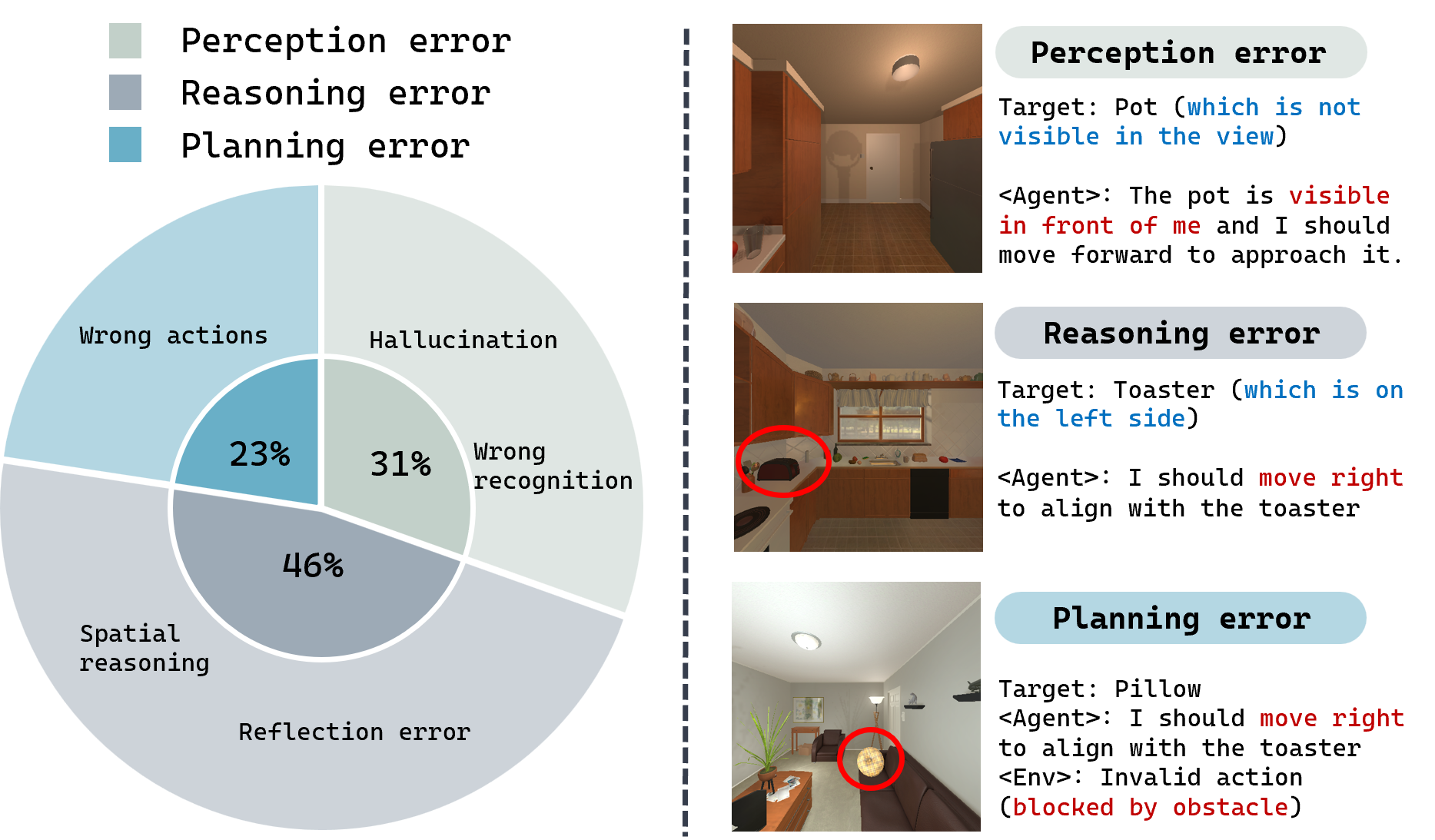}
   \caption{\textbf{The key motivations.} \textbf{Left:} Analysis for different error types from VLA models. \textbf{Right:} Examples of different error types in the navigation task.}
   \label{fig:motivation}
\end{figure}

In Vision-and-Language Navigation (VLN) ~\cite{yin2024sgnav,chen2021hamt,an2023bevbert}, an agent is required to continuously interact with the environment based on the natural language instructions and finally reach a target location. In each step, the environment provides the agent with visual state of the current scene. The agent needs to analyze the current view in combination with the instruction and output an action plan for navigation, which makes the VLN a typical Vision-Language-Action (VLA) problem. However, as \cref{fig:motivation} shows, a VLA model often encounter errors at the levels of perception, reasoning and planning, which also limit its performance on the VLN task ~\cite{yang2025embodiedbench}. Visual hallucination is a typical kind of perception error, where the agent may incorrectly claim to see an object that is not actually in its field of view, or fail to recognize an object that is present. Reasoning errors mainly include incorrect understandings of spatial relationships and improper reflections on environmental feedback. Planning errors mainly include invalid or infeasible actions, such as moving towards obstacles.

Among the aforementioned error types, visual hallucination in LVLMs is a major culprit behind their suboptimal performance in VLA tasks ~\cite{zhang2026vlm4vla}. Some of the existing works attempt to add additional markers to the input images, such as obstacle masks or potential path indicators, to introduce more explicit visual cues and improve the perception ability of the agent ~\cite{pmlr-v288-goetting25a,wang2024vlm}. Such technique is also known as visual prompt (VP). However, existing works lack research on effectively utilizing multi-view inputs through VP, and systematic analysis on how different VP types can complement each other is also missing.
% However, existing visual prompt approaches for navigation tasks are rarely designed specifically to reduce visual hallucination or to enhance the model’s spatial understanding.
% Some other works try to mimic human navigation behavior by having the VLN agent first predict the image after taking an action, and then use information from both the predicted and current scenes to decide the next action ~\cite{li2023vln-sig,perincherry2025visual}. Essentially, these methods aim to incorporate information about the consequences of actions into the decision-making process, which might to some extent enhance the reasoning ability of the VLN agent. However, since such approach relies heavily on images predicted by the model itself, inaccurate predictions are likely to cause unexpected negative effects. 
% In addition to the aforementioned methods, 
Besides, to mitigate errors in reasoning and planning, Supervised Fine-Tuning (SFT) and Reinforcement Fine-Tuning (RFT) are often applied to enhance VLN agents ~\cite{liu2025nav,VLNR1,gao2025octonav}. However, traditional RFT approaches only utilize the sparse outcome reward, while process-based RFT methods often calculate action advantage with respect to the identical anchor state ~\cite{feng2025group}. When the anchor state is hard to define, the computation becomes inefficient and the scalability is limited. Overcoming the anchor-state limitation and developing an efficient, stable process-based RFT algorithm for navigation remains challenging.

To address above challenges, we introduce a novel framework named \textbf{SeeNav-Agent}. First, to reduce the hallucinations of the VLN agent and enhance its ability of spatial understanding, we introduce a dual-view visual prompt method. By introducing VP, the probability of perception and reasoning errors can be reduced. In addition, by using an action projection mechanism within VP, we transform the VLA task into a Visual Question Answering (VQA) task, thereby reducing the difficulty of the navigation. Subsequently, we further enhance the agent’s reasoning and planning abilities through RFT. During RFT, a verifiable process reward for each navigation step is designed. We further propose the \textbf{S}tep \textbf{R}eward \textbf{G}roup \textbf{P}olicy \textbf{O}ptimization (\textbf{SRGPO}) algorithm, which utilizes the process reward signals by randomly grouping navigation steps and computing relative advantages.

% We evaluate the proposed SeeNav-Agent on the EmbodiedBench Navigation dataset ~\cite{yang2025embodiedbench}. xxx (add performance)

The main contribution of this work is summarized as:
\begin{itemize}
    \item We propose a dual-view visual prompt technique for navigation task, which significantly reduces the visual hallucinations and improves the spatial understanding capability of the VLN Agent with a zero-shot setting.
    \item We design a novel RFT algorithm, namely SRGPO, which efficiently integrates the step-level rewards of the navigation process, thereby further enhancing the agent's reasoning and planning abilities.
    % \item We achieve state-of-the-art success rates on navigation tasks on the EmbodiedBench Navigation dataset, and validate the effectiveness of each component of our method through extensive comparative and ablation experiments.
    \item Experiments on Embench-Nav and VLN-CE show that GPT4.1+VP surpasses the SOTA closed-source model by 20 pp (zero-shot) on Embench-Nav and by 11.1 pp on VLN-CE, and Qwen2.5-VL-3B-Instruct+VP+SRGPO exceeds the previous SOTA closed-source model by 5.6 pp on Embench-Nav and improves over the baseline by 8.9 pp on VLN-CE. Moreover, SRGPO shows superior convergence, stability, and generalization.
\end{itemize}

% ==========================================================================
% Section: Related Work  (from sec/2_related_work.tex of the ECCV source, verbatim)
% ==========================================================================
\section{Related Work}
\label{sec:formatting}
\noindent\textbf{Vision-and-Language Navigation (VLN)}. VLN is a type of multimodal embodied VLA task. Traditional VLN pipelines treat image perception, map node segmentation, reasoning and action planning as independent modules, and the agent usually moves in a discrete environment ~\cite{cao2024cognav,majumdar2020improving,rxr,yin2025unigoal}. Recent research based-on LVLMs has provided a unified solution for VLN tasks. The LVLMs-based VLN agents use powerful multimodal models as the backbone, and fuse perception, reasoning, and planning in an end-to-end manner for continuous navigation ~\cite{zhang2025mem2ego,liu2025nav}. These works either use closed-source models directly for navigation ~\cite{pmlr-v288-goetting25a} or fine-tune open-source ones ~\cite{VLNR1,gao2025octonav}.

\noindent\textbf{Visual Prompt (VP) for VLN Agents}. VP methods add visual cues such as bounding boxes, color blocks, and numerical marks to the input image so that LVLMs can better handle visual tasks ~\cite{shtedritski2023does,yang2023setofmark,nasiriany2024pivot}. Since LVLM pre-training corpora are dominated by VQA data, LVLMs are inherently adept at VQA, and VP allows planning-style VLA tasks like VLN to be reformulated as zero-shot VQA. PIVOT ~\cite{nasiriany2024pivot} projects candidate actions as numbered arrows and iteratively optimizes the action distribution via LVLM selection. VLMnav ~\cite{pmlr-v288-goetting25a} annotates walkable-area masks and farthest-reachable arrows, and similarly lets the agent select an optimal action from projected candidates. 
% Finally, an additional LVLM is used to determine the termination of navigation. 
% However, the aforementioned VP method is not specifically designed to alleviate visual hallucinations or enhance the model's ability to understand spatial relationships. In this work, we have effectively addressed the shortcomings of the VLN agent in terms of hallucinations and spatial relationship understanding by carefully designing the relevant components of VP and introducing multi-view inputs.
However, the aforementioned works lack systemic research on how to effectively utilize multi-view inputs through VP and how to coordinate different VP modules for better performance.

\noindent\textbf{Reinforcement Fine-Tuning (RFT) for VLN Agents}. 
RFT is a common post-training method for VLN agents. Proximal Policy Optimization (PPO) \cite{zhai2024finetuning} and Group Relative Policy Optimization (GRPO) ~\cite{VLNR1,gao2025octonav,li2025compassnav} are widely used RFT frameworks, but classic GRPO relies only on sparse outcome rewards, which is inefficient for long-horizon navigation. Process rewards have been shown to effectively enhance RFT ~\cite{wang2025spa,wang2023math,chan2024dense}. GiGPO ~\cite{feng2025group} incorporates step-level rewards by rolling out trajectories from the same initial state and grouping identical states. However, identical states are hard to define in continuous navigation, and achieving sufficient group sizes requires many rollouts, causing a heavy computational load. 
% VAGEN ~\cite{wang2025vagen} applies turn-level advantage estimation to PPO. Its limitation lies in the need to provide textual descriptions for each state, thereby using an Large Language Model (LLM)-based world model to score the reasoning quality of each turn and obtain turn-level reward signals. Additionally, due to the adoption of the PPO architecture, the computational resources it requires will exceed those of GRPO-like methods. 
In this work, a robust and verifiable process reward for navigation tasks is designed for obtaining step-level reward signals. Meanwhile, we improve the GRPO framework and propose SRGPO, an RFT algorithm based on random step grouping, which efficiently addresses the issue in GiGPO that requires grouping of identical states.

% ==========================================================================
% Section: Methodology  (from sec/3_methodology.tex of the ECCV source, verbatim)
% ==========================================================================
\section{Methodology}
\subsection{Framework and Problem Formulation}
\begin{figure*}[htbp]
  \centering
   \includegraphics[width=0.98\textwidth]{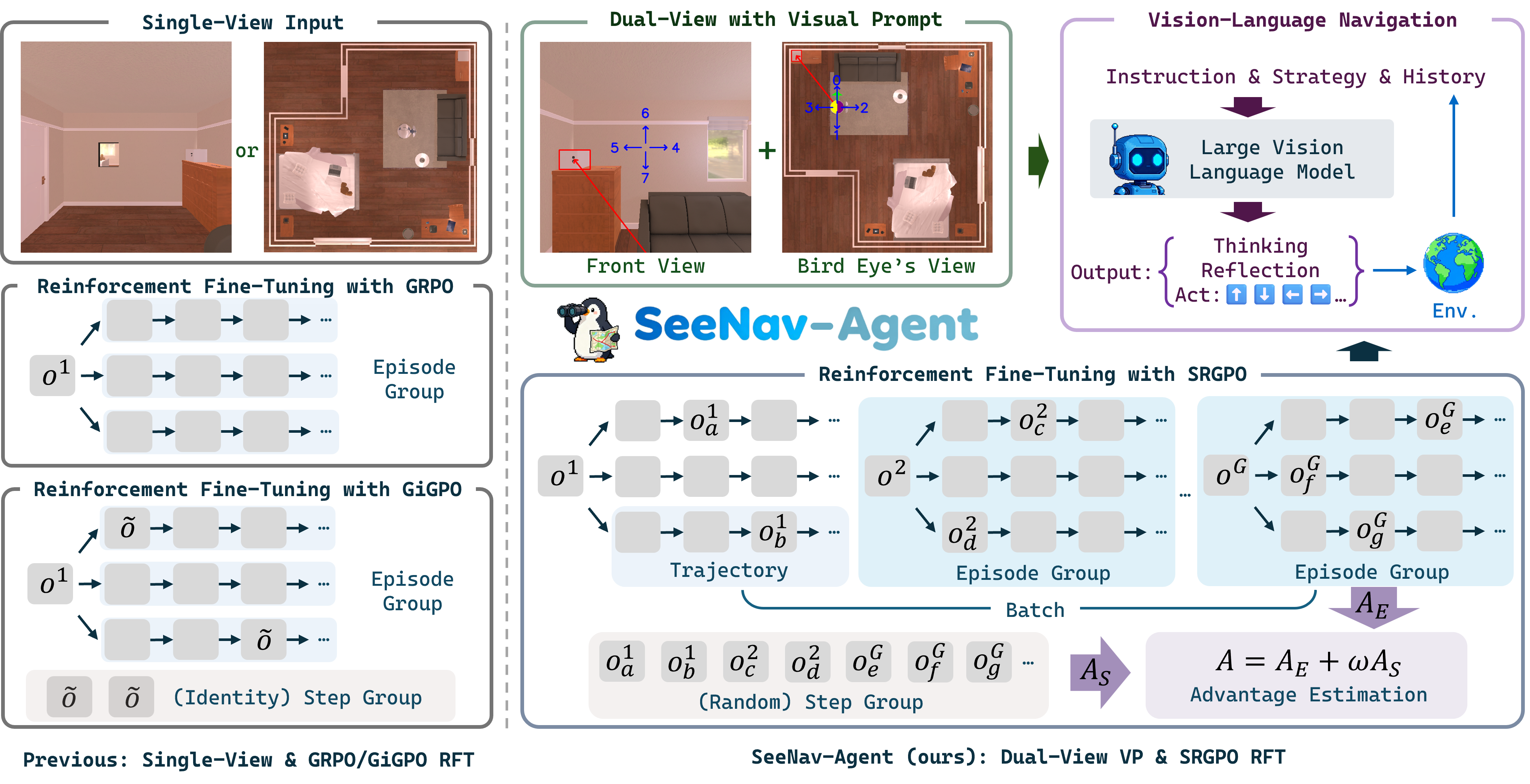}
   \caption{\textbf{Overview of SeeNav-Agent.} Different from previous VLN works tend to use single view image as input and use method like GRPO or GiGPO for RFT, our SeeNav-Agent designs a dual-view input with visual prompt to enhance the visual module in a zero-shot manner, and proposes SRGPO to efficiently introduce process reward signals by randomly grouping steps. The upper-middle schematic shows the dual-view visual prompt, with FV on the left and BEV on the right.}
   \label{fig:framework}
\end{figure*}

In this section, we introduce a novel LVLM-based embodied navigation framework named SeeNav-Agent. As shown in \cref{fig:framework}, SeeNav-Agent includes a specifically designed dual-view visual prompt technique for the input to reduce perception hallucinations and to enhance the spatial understanding capability of the agent. During RFT, to overcome the limitations of previous RFT algorithms like GRPO and GiGPO, we propose a novel algorithm named SRGPO that can efficiently utilize both episode-level and step-level navigation rewards in the advantage estimation stage.

We first formalize the LVLM-based VLN. The input state of the VLN agent at step $t$ is $\mathbf{O}_t=\{\mathbf{I},\mathbf{S},\mathbf{V}_t,\mathbf{H}_t\}$, where $\mathbf{I}$ is the human instruction; $\mathbf{S}$ represents the action strategy given by humans, which is embedded in the input context; $\mathbf{V}_t$ denotes the environmental image observed by the agent at step $t$ and $\mathbf{H}_t$ is the action interaction history before step $t$. In this work, we use an action interaction history with a time window as $\mathbf{H}_t=(\mathbf{a}_{t-T_H},\mathbf{f}_{t-T_H},\mathbf{a}_{t-T_H+1},\mathbf{f}_{t-T_H+1},\dots,\mathbf{a}_{t-1},\mathbf{f}_{t-1})$, where $\mathbf{a}_t$ refers to the action at step $t$, and $\mathbf{f}_t$ denotes the feedback from the environment for this action. Both $\mathbf{a}_t$ and $\mathbf{f}_t$ are presented in text form. $T_H$ represents the time window for action history. The output of the agent in each step is $(\mathbf{c}_t,\mathbf{a}_t)$, where $\mathbf{c}_t$ represents the analysis and thinking process about the current state, and $\mathbf{a}_t\in\mathcal{A}$ is the action decision chosen from the action space $\mathcal{A}$. The LVLM-based VLN agent is parameterized as a policy network $\pi_\theta(\mathbf{a}_t|\mathbf{O}_t)$ with learnable parameters $\theta$.

\subsection{Dual-View Visual Prompt for VLN Agent}
\label{sec:vp}
Most existing VLN works tend to use only a single perspective as the input for the agent, such as a front view (FV) or a bird's-eye view (BEV). However, due to the poor ability of LVLM itself to understand spatial information like depth of field, the spatial relationships of objects in the FV are not as clear as those in the BEV for LVLM. But the problem with relying solely on the BEV is that the shapes of objects in it are often different from common representations, so an LVLM may fail to recognize a target object in the BEV. Therefore, we consider enabling the agent to see both FV and BEV simultaneously. This dual-view (DV) input can enrich the visual information, but it also increases the difficulty for the agent to process and understand the images. If utilized inappropriately, the performance of the agent may not improve but instead decline ~\cite{yang2025embodiedbench}. Consequently, we design a series of VP modules as follows to help the agent better understand the visual information. An example of these VP modules is shown in the upper-middle of \cref{fig:framework}.

\noindent\textbf{Bounding box}. For navigation tasks such as ``moving toward a target object'', incorrect judgment of the target's existence is one of the most common visual hallucinations. The agent may claim to see an object that does not exist in the view, or ignore a target that is clearly within its field of view, leading to decision-making errors. We enhance the existence information of the target by drawing bounding boxes (BB) around it in both FV and BEV, and also add relevant prompts to reduce hallucinations in judging the target existence.

\noindent\textbf{Navigation line}. Inspired by the guidance trajectories in the minimaps of open-world games, we add navigation line (NL) arrows from the agent to the target. Since FV presents the first-person perspective of the agent, the starting point of the NL is set at the midpoint of the bottom edge. In BEV, however, the starting point of the NL is the agent's center point. NL can enhance the agent's understanding of the relative position and distance between itself and the target.

\noindent\textbf{Agent marker}. In navigation, enabling the agent to understand its position in space helps it plan the next action. Therefore, we use the agent marker (AM) represented by a circle and an arrow in the BEV to indicate the position and orientation of the agent. Additionally, since the agent often makes mistakes in understanding left-right relationships, we mark the left and right sides of the circle with different colors (yellow for left, purple for right, and a green arrow for the agent's front), which enhances its comprehension of the relative positional relationship between itself and the target.

\noindent\textbf{Action projection}. The main actions in navigation include movement in various directions and perspective rotation, which can be projected onto images. Through action projection (AP), the agent can partly observe the consequences of executing a specific action. Besides, it can also convert a planning problem into VQA of selecting the optimal action on the image, reducing the difficulty of navigation. We draw movement actions in the BEV and draw rotation actions in the FV with arrows and action IDs (0--7 in the Embench-Nav illustration), thus projecting the action space.

\noindent\textbf{View alignment}. In BEV, the agent can face to various directions due to the randomness of the BEV orientation, which may lead the agent to mistakenly understand the left-right and front-back relationships between itself and objects (such errors may still occur even with the AM). Therefore, we rotate the BEV according to the agent's orientation, ensuring that the agent always faces forward in the BEV, which means its viewing direction is consistent with FV.

% The modules mentioned above make up the visual prompt scheme used in this work and the upper-middle schematic of \cref{fig:framework} shows the application effect of the dual-view visual prompt that incorporates all these components.
Actually, VP does not require every cue from privileged state: only the target bounding box and the BEV need external perception. Navigation lines and action projections are derived from these plus the agent pose.
In this work we directly use the environment’s intrinsic information to obtain BEV and bounding boxes, which is a natural choice for simulation, games, and other environments where the full state is accessible. For settings that are closer to the real world, one can rely on additional auxiliary models or tools to obtain the required information. For example, works like BEVFormer ~\cite{li2024bevformer} and CorrBEV ~\cite{Xue_2025_CVPR} focus on BEV modeling, and there is extensive research on vision‑based object detection ~\cite{zhang2022glipv2, zhang2022dino, wang2024yolov10}.
Our proposed VP approach is a visual interface orthogonal to and decoupled from its signal source.

\subsection{Step Reward Group Policy Optimization}
\label{sec:srgpo}
% 1. While VP can mitigate perception error, 但是进一步解决planning error来需要引入RFT来解决。
% 2. GRPO用outcome reward作为policy的学习信号，但是GRPO suffers from sparse learning signal. 
% 3. Existing works like GiGPO propose step-level optimization method.
% 4. However, the definition of identical states is relatively strict.
% 5. To solve this challenge, we propose SRGPO.
While VP can mitigate perception error, further addressing reasoning and planning errors still requires post-training with RFT. Classic GRPO~\cite{shao2024deepseekmath} learns from the outcome reward, however, for agentic problems like VLN that require long-sequence interactions with the environment, GRPO suffers from sparse reward signals, and how to assign credit to each step becomes a challenging problem. Existing work like GiGPO~\cite{feng2025group} proposes a step-level optimization method based on the identical anchor state. However, the definition of identical states is relatively strict, and for problems where the anchor state is hard to define, the computational burden will significantly increase. To solve this challenge, we propose SRGPO with verifiable process rewards, which leverages the unique properties of navigation tasks. The core insight of SRGPO is that if we can define a process reward that is almost independent of the state, a flexible grouping of steps and advantage estimation can be achieved. SRGPO overcomes the drawbacks of GRPO, which fails to consider process rewards, and the strict prerequisite of GiGPO, which requires grouping identical states, thereby achieving efficient utilization of process reward signals during the navigation process.

\noindent\textbf{Verifiable process reward for navigation}. For navigation, whether the final result is successful or not is a clear outcome reward. However, for each navigation step, the quality of each action can also be judged. For example, an action that brings the agent closer to the target, or pulls the target back into the FV, can be considered good. We denote the agent position at time step $t$ as $p_t$, and the target object as $g$. The visible object set from FV at time step $t$ is defined as $\mathcal{F}_t$.
%Based on this insight, 
We define the verifiable process reward (VPR) $R^s_t$ for navigation as follows:
\begin{equation}
\begin{aligned}
    R^s_t &= R^s_{t,\text{base}} -\lambda_{\text{valid}} \cdot R^s_{t, \text{valid}}, \label{eq:process_reward}\\
    R^s_{t,\text{base}} &=
    \begin{cases}
    1, & \text{if }\mathrm{dist}(p_{t+1}, g)<\mathrm{dist}(p_{t},g)\ \text{or}\ \mathbf{1}_{\{g \in \mathcal{F}_{t+1}\}}>\mathbf{1}_{\{g \in \mathcal{F}_{t}\}}, \\
    0, & \text{otherwise},
    \end{cases}\\
    R^s_{t,\text{valid}} &= \mathbf{1}_{\{\mathbf{a}_t\notin \mathcal{A}_t^{\mathrm{feasible}}\}}.
\end{aligned}
\end{equation}
where $\text{dist}(p_t,g)$ is the distance between the agent and the target at step $t$, and $g \in \mathcal{F}_t$ means the target is in the FV at step $t$. Since $\mathbf{a}_t$ is selected from $\mathbf{O}_t$ and executed to reach $p_{t+1}$, $R^s_t$ evaluates the transition caused by $\mathbf{a}_t$. $R^s_{t,\text{valid}}$ describes the penalty for invalid actions, where $\mathcal{A}_t^{\mathrm{feasible}}\subseteq\mathcal{A}$ is the set of actions that can be successfully executed at step $t$ (\eg, not blocked by obstacles), as reported by the environment feedback $\mathbf{f}_t$. 
The design of the process reward relies solely on the distance changes between the agent and the target for two consecutive steps, as well as the changes in the visibility of the target in the FV. 
The key insight of this reward design is that it is independent of the specific environmental state: a reward of \mbox{$1$} always denotes verifiable local progress toward the target and \mbox{$0$} the absence of it. 
Therefore, the quality of actions can be compared under any two different states, which is the prerequisite for the design of SRGPO.

\noindent\textbf{Episode-level advantages}.
We first introduce the episode-level advantage of vanilla GRPO. We roll out $N$ trajectories $\{\boldsymbol{\tau}_i\}^{N}_{i=1}$ using the agent policy $\pi_{\theta_\text{old}}$ for one task, where each trajectory $\boldsymbol{\tau}_i=\{(\mathbf{O}_{i,0},\mathbf{a}_{i,0}),(\mathbf{O}_{i,1},\mathbf{a}_{i,1}),\dots,(\mathbf{O}_{i,T_i-1},\mathbf{a}_{i,T_i-1})\}$. The outcome reward for each trajectory is defined as $R(\boldsymbol{\tau}_i)$. For navigation, the reward is 1 if the agent completes the task and 0 otherwise. Then, the episode-level group and the corresponding normalized advantage are:
\begin{align}
  G_E&=\{(\boldsymbol{\tau}_1,R(\boldsymbol{\tau}_1)),\dots,(\boldsymbol{\tau}_{N},R(\boldsymbol{\tau}_{N}))\},
  \label{eq:G_E}\\
  A_E(\boldsymbol{\tau}_i)&=\frac{R(\boldsymbol{\tau}_i)-\mathrm{mean}(\{R(\boldsymbol{\tau}_j)\}^{N}_{j=1})}{\mathrm{std}(\{R(\boldsymbol{\tau}_j)\}^{N}_{j=1})}.
  \label{eq:A_E}
\end{align}

\noindent\textbf{Step-level advantages}. The episode-level reward signals are relatively sparse, while RFT tends to benefit from denser reward signals. Therefore, we use verifiable process reward signals to calculate step-level advantages. Similar to GiGPO, calculating step-level advantages requires grouping steps first. However, unlike GiGPO, since the verifiable process rewards designed in \cref{eq:process_reward} do not depend on specific states, we do not need to define the anchor states which describe state identity as in GiGPO. The core idea is that, through the definition of verifiable process rewards, we discard the prerequisite that ``to compare the quality of two actions, it must be done under the same state''. Therefore, when calculating step-level advantages, we can change the concept of groups in the episode-level advantage calculation and instead randomly sample steps from the \textit{entire batch} to form step groups. Assuming one batch contains $B$ tasks, we have $N\times B$ trajectories, and the step-level groups with size $N_S$ can be defined as:
\begin{equation}
  G_S = \bigl\{(\mathbf{c}^{(i)}_t,\mathbf{a}^{(i)}_t,R^{s(i)}_t)\bigr\}_{N_S},\quad t\in\{0,\dots,T_i-1\},\; i\in\{1,\dots, N\times B\},
\label{eq:G_S}
\end{equation}
where $i$ and $t$ are randomly sampled from their valid range, and $T_i$ represents the episode length of the $i$-th trajectory. The step-level advantage for a step within the group is then:
\begin{equation}
  A_S(\mathbf{c}^{(i)}_t,\mathbf{a}^{(i)}_t)
  =\frac{R^{s(i)}_t - \mathrm{mean}_{G_S}(R^{s})}{\mathrm{std}_{G_S}(R^{s})},
\label{eq:A_S}
\end{equation}
where $\mathrm{mean}_{G_S}(R^{s})$ and $\mathrm{std}_{G_S}(R^{s})$ are the mean and standard deviation of process rewards in $G_S$.

\noindent\textbf{SRGPO training objective}. The two levels of advantages are combined into a unified advantage for both step-level and trajectory-level credit assignment. With balance coefficient $\omega$, the combined advantage and the SRGPO objective are:
\begin{align}
  A_{i,t}&=A_E(\boldsymbol{\tau}_i) + \omega \cdot A_S(\mathbf{c}^{(i)}_t,\mathbf{a}^{(i)}_t), \label{eq:A}\\
  \mathcal{J}_{\text{SRGPO}}(\theta)&=\mathbb{E}_{{\{\boldsymbol{\tau}_i\}^{N}_{i=1}\sim\pi_{\theta_\text{old}}}}
  \Bigl[\tfrac{1}{N}{\textstyle\sum_{i=1}^{N}}\tfrac{1}{T_i}{\textstyle\sum_{t=0}^{T_i-1}}\Bigl(\min\bigl(\rho_{i,t}(\theta)A_{i,t},\mathrm{clip}(\rho_{i,t}(\theta),1\pm\epsilon)A_{i,t}\bigr) \nonumber\\
  &\qquad-\beta\,\mathbb{D}_{\text{KL}}\bigl(\pi_{\theta}(\cdot|\mathbf{O}_{i,t})\,\|\,\pi_{\text{ref}}(\cdot|\mathbf{O}_{i,t})\bigr)\Bigr)\Bigr],
  \label{eq:SRGPO_obj}
\end{align}
where $\rho_{i,t}(\theta)=\frac
{\pi_{\theta}(\mathbf{c}^{(i)}_t,\mathbf{a}^{(i)}_t|\mathbf{O}_{i,t})}
{\pi_{\theta_{\text{old}}}(\mathbf{c}^{(i)}_t,\mathbf{a}^{(i)}_t|\mathbf{O}_{i,t})}$ is the importance sampling ratio and $\beta$ is the KL penalty coefficient. Both the importance ratio and the KL penalty are computed over the complete generated response of each step, \ie, the reasoning tokens $\mathbf{c}_t$ together with the action tokens $\mathbf{a}_t$, and the same advantage $A_{i,t}$ is shared by all tokens of this response.
% where $\rho_{i,t}(\theta)=\frac
% {\pi_{\theta}(\mathbf{c}^{(i)}_t,\mathbf{a}^{(i)}_t|\mathbf{o}^{(i)})}
% {\pi_{\theta_{\text{old}}}(\mathbf{c}^{(i)}_t,\mathbf{a}^{(i)}_t|\mathbf{o}^{(i)})}$ is the importance sampling ratio and $\beta$ is the KL penalty coefficient.

% ==========================================================================
% Section: Experiments  (from sec/4_experiments.tex of the ECCV source, verbatim)
% Layout adjustments: tab:main_results is upgraded to a two-column-spanning
% \begin{table*} with \resizebox{0.95\textwidth}{!}; tab:ablation_vp keeps
% single-column \begin{table} but uses \resizebox{\columnwidth}{!}; figure*
% widths use \textwidth; figure paths point to Figures/figs/.
% ==========================================================================
\section{Experiments}

\subsection{Experimental Setup}
\label{sec:setup}
\noindent\textbf{Benchmark}.
We evaluate SeeNav-Agent on Embench-Nav~\cite{yang2025embodiedbench} and VLN-CE~\cite{krantz_vlnce_2020}. Embench-Nav contains 60 AI2-THOR~\cite{kolve2017ai2} in-room tasks; VLN-CE is a Habitat-Sim~\cite{puig2023habitat} continuous indoor benchmark. Action spaces, scene parameters, and data splits are given in \cref{sec:Experimental-Details}.

\noindent\textbf{Baselines and metrics}.
We compare against (i) closed-source LVLMs, (ii) open-source LVLMs, and (iii) post-training methods (SFT, GRPO, GiGPO), using Qwen2.5-VL-3B-Instruct~\cite{bai2025qwen2} as the backbone for RFT. The main metric is success rate (SR); on VLN-CE we also report SPL, NE, and a comparison with VLN-R1~\cite{VLNR1}.

\noindent\textbf{Implementation details}.
Hyper-parameters are shared across both benchmarks: $\lambda_{\text{valid}}=0.1$, $\omega=0.5$ in \cref{eq:A}, $N=4$, $N_S=16$, and $T_H=5$. VP is kept throughout RFT, and dual-view inputs concatenate FV (left) and BEV (right). Other setup details are in the appendix.

\subsection{Main Results}
% To demonstrate the strong performance of the SeeNav-Agent on navigation tasks, we compare the proposed method with numerous open-source and closed-source models. 
The main results are shown in \cref{tab:main_results}. Compared with the close-source models, our VP technique improves the SR of GPT4.1 by 21.7 pp and surpasses the best Claude-3.5-Sonnet by 20.0 pp on Embench-Nav, and achieves an 11.1 pp improvement of SR on VLN-CE. Compared with the open-source models, our method achieves a 55.6 pp improvement on the base model Qwen2.5-VL-3B-Ins and surpasses the best InternVL3-78B by 5.6 pp on Embench-Nav, and beats the baseline VLN-R1 by 8.9 pp of SR on VLN-CE. Moreover, our SRGPO significantly outperforms post-training baselines including SFT, GRPO and GiGPO, confirming the effectiveness of SeeNav-Agent.

% The results are shown in \cref{tab:main_results}. As shown by the results, when VP is added to the advanced closed-source model GPT4.1, its navigation performance surpasses that of the best open-source and closed-source models by about 20 pp. Furthermore, when SRGPO is used to perform RFT on the Qwen2.5-VL-3B model, its performance exceeds that of the best models by about 5.6 pp. These results fully demonstrate the effectiveness of the proposed method. \zichuan{should be improved}
\begin{table*}[t]
  \caption{\textbf{Main Results.} RFT cases are trained with i.d. settings}
  \label{tab:main_results}
  \centering
  \scriptsize
  \setlength{\tabcolsep}{3pt}
  \renewcommand{\arraystretch}{0.9}
  \begin{adjustbox}{max width=\linewidth}
  \begin{tabular}{lcccc}
    \toprule
     & \textbf{Embench-Nav} & \multicolumn{3}{c}{\textbf{VLN-CE(R2R)}} \\
    \cmidrule(lr){2-2}\cmidrule(lr){3-5}
    \textbf{Method} & \textbf{SR}\textuparrow & \textbf{SR}\textuparrow & \textbf{SPL}\textuparrow & \textbf{NE}\textdownarrow \\
    \midrule
    \multicolumn{5}{l}{\textit{Close-Source LVLMs}} \\
    % Claude-3.7-Sonnet & 0.500 & & & \\
    \underline{Claude-3.5-Sonnet} & \underline{0.667} & & & \\
    % GPT-4o & 0.550 & & &\\
    % GPT-4o-mini & 0.317 & & &\\
    \textcolor{RoyalBlue}{GPT4.1} & \textcolor{RoyalBlue}{0.650} & \textcolor{RoyalBlue}{0.141} & \textcolor{RoyalBlue}{0.117} & \textcolor{RoyalBlue}{8.8} \\
    % Gemini-1.5-Pro & 0.233 & & & \\
    Gemini-2.0-flash & 0.633 & & & \\
    % Gemini-1.5-flash & 0.567 \\
    Qwen-VL-Max & 0.500 & & & \\
    \midrule
    \multicolumn{5}{l}{\textit{Open-Source LVLMs}} \\
    % Llama-3.2-90B-Vision-Ins & 0.233 & & & \\
    % Llama-3.2-11B-Vision-Ins & 0.483 & & & \\
    % InternVL2 5-78B & 0.367 \\
    % InternVL2 5-38B & 0.350 \\
    % InternVL2 5-8B & 0.350 \\
    \underline{InternVL3-78B} & \underline{0.667} & & & \\
    % InternVL3-38B & 0.550 \\
    % InternVL3-8B & 0.383 & & & \\
    % Qwen2-VL-72B-Ins & 0.267 \\
    % Qwen2-VL-7B-Ins & 0.267 \\
    Qwen2.5-VL-72B-Ins & 0.467 & & & \\
    Qwen2.5-VL-7B-Ins & 0.200 & & & \\
    \textcolor{ForestGreen}{Qwen2.5-VL-3B-Ins} & \textcolor{ForestGreen}{0.167} & \textcolor{ForestGreen}{0.112} & \textcolor{ForestGreen}{0.083} & \textcolor{ForestGreen}{9.7} \\
    % Ovis2-34B & 0.633 & & & \\
    % Ovis2-16B & 0.600 \\
    % gemma-3-27b-it & 0.533 & & & \\
    % gemma-3-12b-it & 0.383 \\
    \midrule
    % \multicolumn{2}{c}{\textit{SFT/RFT baselines (backbone: \textcolor{ForestGreen}{Qwen2.5-VL-3B-Ins})}} \\
    \multicolumn{5}{l}{\textit{Post-Train baselines (backbone: Qwen2.5-VL-3B-Ins)}} \\
    VP+SFT & 0.367 & & & \\
    VP+GRPO & 0.408$\pm$0.008 & 0.230$\pm$0.067 & 0.191$\pm$0.072 & 8.6$\pm$2.4 \\
    VP+GiGPO (vanilla) & 0.294$\pm$0.028 & 0.226$\pm$0.070 & 0.190$\pm$0.067 & 8.6$\pm$2.6\\
    VP+GiGPO (w. VPR) & 0.572$\pm$0.080 & 0.281$\pm$0.042 & 0.220$\pm$0.041 & 7.8$\pm$1.9 \\
    VLN-R1 \textit{\scriptsize (Qwen2-VL-7B)}~\cite{VLNR1} & - & 0.302 & 0.218 & 7.0 \\
    \midrule
    \multicolumn{5}{l}{\textit{SeeNav-Agent (ours)}} \\
    % Qwen2.5-VL-3B-Ins+VP+SRGPO(sg8) & 0.411$\pm$0.103 \\
    % \textcolor{ForestGreen}{\textbf{Qwen2.5-VL-3B-Ins+VP+SRGPO}}  & \textcolor{ForestGreen}{\textbf{0.723$\pm$0.008}} \\
    % Qwen2.5-VL-3B-Ins+VP+SFT+SRGPO & 0.433 \\
    % VP+SFT+SRGPO & 0.433 \\
    VP-only \textit{\scriptsize (Qwen2.5-VL-3B)} & 0.283 & 0.184 & 0.165 & 7.8 \\
    SRGPO-only \textit{\scriptsize (Qwen2.5-VL-3B)} & 0.382$\pm$0.090 & 0.246$\pm$0.102 & 0.211$\pm$0.081 & 8.3$\pm$2.5 \\
    \textcolor{ForestGreen}{\textbf{VP+SRGPO}} \textit{\scriptsize (Qwen2.5-VL-3B)} & \textcolor{ForestGreen}{\textbf{0.723$\pm$0.008} (\textuparrow{0.556})} & \textcolor{ForestGreen}{\textbf{0.391$\pm$0.021}(\textuparrow{0.279})} & \textcolor{ForestGreen}{\textbf{0.327$\pm$0.020}(\textuparrow{0.244})} & \textcolor{ForestGreen}{\textbf{5.1$\pm$1.8}(\textdownarrow{4.6})}\\
    \textcolor{RoyalBlue}{\textbf{GPT4.1+VP}} \textit{\scriptsize (GPT4.1)} & \textcolor{RoyalBlue}{\textbf{0.867 (\textuparrow{0.217})}} & \textcolor{RoyalBlue}{0.252(\textuparrow{0.111})} & \textcolor{RoyalBlue}{0.208(\textuparrow{0.091})}& \textcolor{RoyalBlue}{7.0(\textdownarrow{1.8})}\\
    \bottomrule
  \end{tabular}
  \end{adjustbox}
\end{table*}

\subsection{Ablation Study for Visual Prompt}
We conducted ablation experiments on each VP modules on the Embench-Nav benchmark and the results are shown in \cref{tab:ablation_vp}. The results show that by introducing VP, a 21.7 pp improvement in navigation success rate can be achieved with a zero-shot manner. From the ablation results, it can be suggested that simply introducing dual-view (DV) input and view alignment (VA) cannot improve the success rate, instead, it may even lead to a decline in the agent's performance. The reason is that it is difficult for the agent to accurately understand and align the spatial information brought by the two different views. Furthermore, the results also indicate that each module in the VP recipe works collectively. It is necessary to include all VP modules simultaneously to achieve the optimal navigation success rate. Among these modules, bounding-boxes (BB), action projection (AP), and view alignment (VA) are relatively more important, which means removing any one of them will lead to a significant drop in the success rate of the navigation task.

% \begin{table}
%   \footnotesize
%   \caption{\textbf{Ablation Study for Visual Prompt.} Avg. steps mean the average number of steps required to complete the task.}
%   \label{tab:ablation_vp}
%   \centering
%   \begin{tabular}{@{}ccc@{}}
%     \toprule
%     \textbf{Method} & \textbf{Succ. rate} & \textbf{Avg. steps} \\
%     \midrule
%     GPT4.1 (FV only) & 0.650 & 14.12 \\
%     DV+VA & 0.450 & 16.00 \\
%     DV+BB+AP+AM+NL & 0.583 & 15.10 \\
%     DV+AP+AM+VA & 0.650 & 14.95 \\
%     DV+BB+AM+NL+VA & 0.683 & 14.32 \\
%     DV+BB+AP+AM+VA & 0.800 & 12.97 \\
%     DV+BB+AP+NL+VA & 0.800 & \textbf{12.88} \\
%     \textbf{DV+BB+AP+AM+NL+VA} & \textbf{0.867} & 13.10 \\
%     \bottomrule
%   \end{tabular}
% \end{table}

\begin{table}[t]
  \scriptsize
  \renewcommand{\arraystretch}{0.9}
  \begin{minipage}[t]{0.49\textwidth}
    \centering
    \caption{\textbf{Ablation Study for Visual Prompt on Embench-Nav.} All combinations are based on GPT4.1 and comb.(A) is the FV only baseline.}
    \label{tab:ablation_vp}
    \setlength{\tabcolsep}{3pt}
    \begin{adjustbox}{max width=\linewidth}
    \begin{tabular}{@{}c cccccc c@{}}
      \toprule
      \textbf{Comb.} & \textbf{DV} & \textbf{BB} & \textbf{AP} & \textbf{AM} & \textbf{NL} & \textbf{VA} & \textbf{SR} \\
      \midrule
      (A) & & & & & & & 0.650 \\
      (B) & \checkmark & & & & & \checkmark & 0.450 \\
      (C) & \checkmark & & \checkmark & \checkmark & & \checkmark & 0.650 \\
      (D) & \checkmark & \checkmark & \checkmark & \checkmark & \checkmark & & 0.583 \\
      (E) & \checkmark & \checkmark & & \checkmark & \checkmark & \checkmark & 0.683 \\
      (F) & \checkmark & \checkmark & \checkmark & \checkmark & & \checkmark & 0.800 \\
      (G) & \checkmark & \checkmark & \checkmark & & \checkmark & \checkmark & 0.800 \\
      (H) & \checkmark & \checkmark & \checkmark & \checkmark & \checkmark & \checkmark & \textbf{0.867} \\
      \bottomrule
    \end{tabular}
    \end{adjustbox}
  \end{minipage}\hfill
  \begin{minipage}[t]{0.48\textwidth}
    \centering
    \caption{\textbf{Grouping Strategy Comparison on Embench-Nav.} Both localized alternatives underperform batch-level random grouping.}
    \label{tab:local-grouping}
    \setlength{\tabcolsep}{4pt}
    \begin{adjustbox}{max width=\linewidth}
    \begin{tabular}{@{}lcc@{}}
      \toprule
      \textbf{Grouping strategy} & \textbf{Scope} & \textbf{SR}\textuparrow \\
      \midrule
      Random step grouping (SRGPO) & batch-level & 0.723$\pm$0.008 \\
      Within-task grouping & task-level & 0.633$\pm$0.031 \\
      Within-trajectory grouping & trajectory-level & 0.583$\pm$0.012 \\
      \bottomrule
    \end{tabular}
    \end{adjustbox}
  \end{minipage}
\end{table}

\begin{figure}[!t]
  \centering
%  \fbox{\rule{0pt}{2in} \rule{0.9\linewidth}{0pt}}
   \includegraphics[width=\columnwidth]{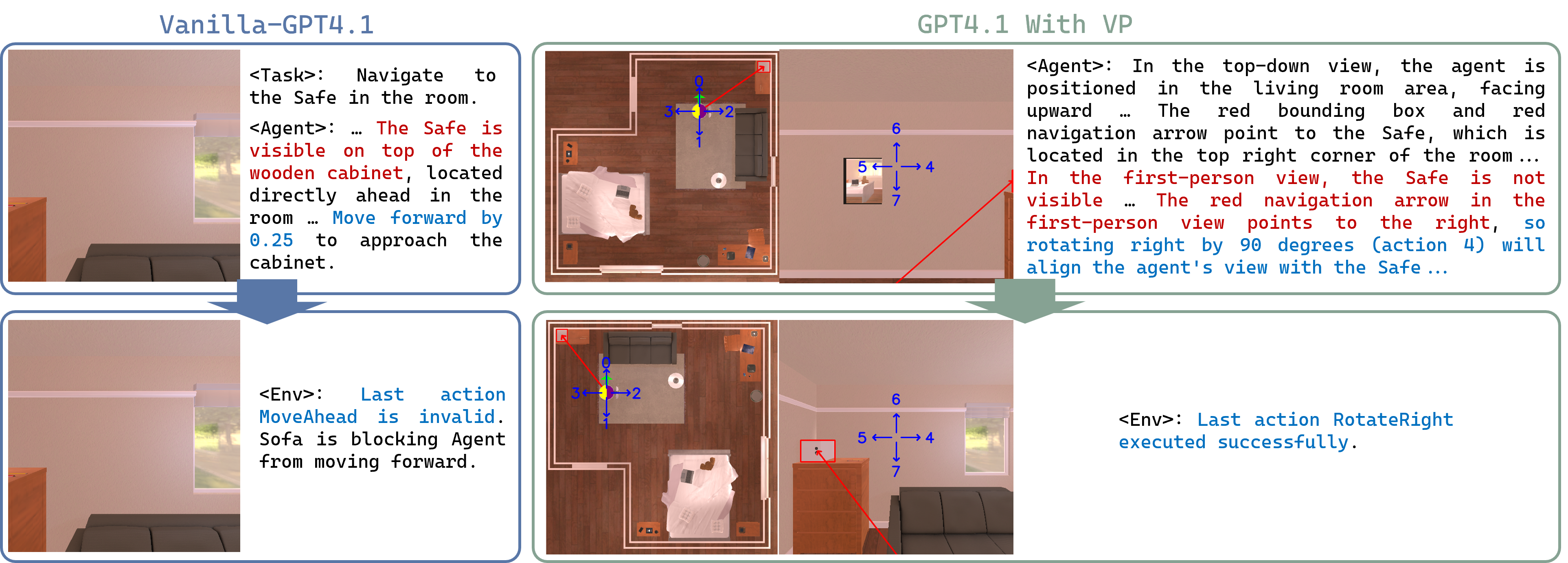}
   \caption{\textbf{Case Comparison of vanilla-GPT 4.1 and GPT4.1 with VP on Embench-Nav.} The bottom part shows the image and text feedback from the environment after taking actions. This case clearly shows that the dual-view VP technique reduces visual hallucination and strengthens the VLN agent's spatial understanding.}
   \label{fig:VP-case}
\end{figure}

\cref{fig:VP-case} shows the differences in the thinking and acting processes of GPT4.1 before and after introducing VP modules when the agent loses sight of the target in FV. In this scenario, the agent is instructed to move to the \textit{Safe}. The reasoning process of the vanilla GPT4.1 model exhibits hallucination, mistakenly believing that the \textit{Safe} is still within its field of view. Consequently, it decides to move forward and fails due to an obstacle. In contrast, after applying VP, the agent can correctly detect that the target is out of sight. Furthermore, with the information from NL, the agent identifies that the target should be on its right side. Finally, among the projected actions, the agent selects action 4 for view rotation and brings the target back into its field of view.

\begin{figure}[!t]
  \centering
%  \fbox{\rule{0pt}{2in} \rule{0.9\linewidth}{0pt}}
   \includegraphics[width=\columnwidth]{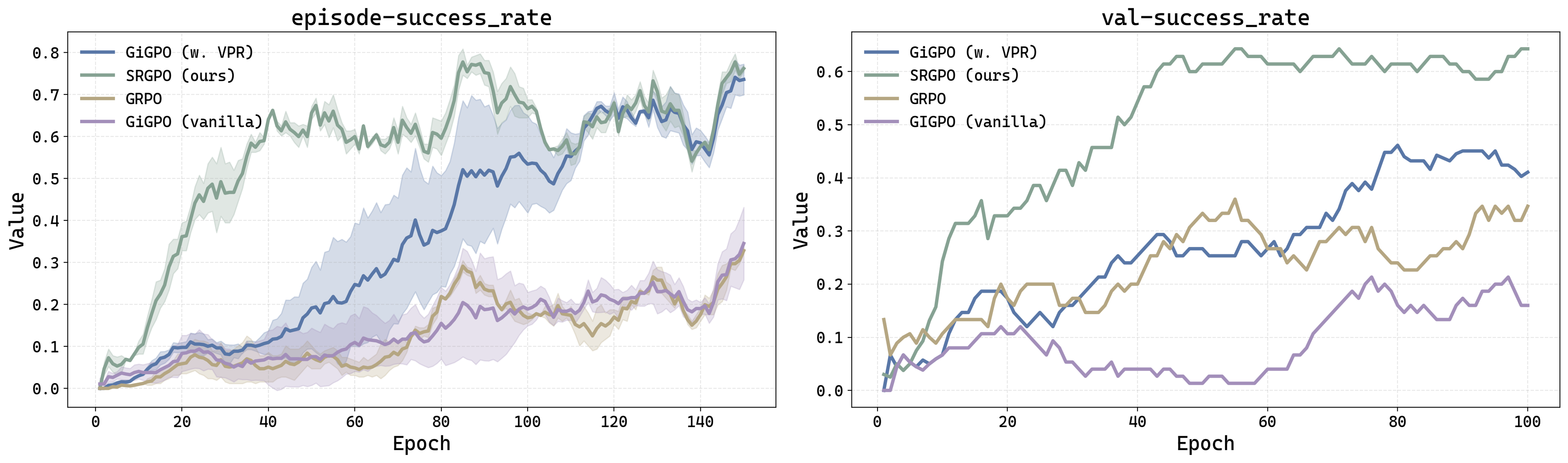}
   \caption{\textbf{i.d and o.o.d Curves on Embench-Nav.} \textbf{Left:} i.d training curves, where the solid line and the shaded region represent the mean and standard deviation obtained from multiple training runs. \textbf{right:} o.o.d. validation curves, where all algorithms are trained on o.o.d. scenes and tested on base scenes of the Embench-Nav.}
   \label{fig:embench_nav_curves}
\end{figure}

\subsection{Effectiveness of SRGPO}
% \begin{figure}[t]
%   \centering
% %  \fbox{\rule{0pt}{2in} \rule{0.9\linewidth}{0pt}}
%    \includegraphics[width=0.95\linewidth]{sec/figs/episode-success_rate-qwen3B.png}
%    \caption{\textbf{Training Process of Qwen2.5-VL-3B-Instruct.} The solid line and the shaded region represent the mean and standard deviation obtained from multiple training runs.}
%    \label{fig:episode-success_rate-qwen3B}
% \end{figure}
The i.d. training curves on Embench-Nav of GRPO, GiGPO and SRGPO are shown in the left part of \cref{fig:embench_nav_curves}. We conduct three runs for each method. For GiGPO, we test both the step reward from the original paper ``VP+GiGPO(vanilla)'' and the verifiable step reward proposed in this work ``VP+GiGPO(w. VPR)''. As can be seen from the figure, ``VP+SRGPO'' significantly outperforms ``VP+GRPO'' and ``VP+GiGPO(vanilla)'' in both convergence speed and final training success rate. Besides, it also demonstrates the best training stability, as evidenced by the very small standard deviation of its training curves. However, when GiGPO adopts the verifiable step reward, it can converge to a success rate close to that of SRGPO, but its convergence speed is still much slower. The reason is that although GiGPO is capable of grouping steps, its grouping efficiency is lower than that of SRGPO due to the constraint of identical states. In other words, the number of steps in each group is relatively small. After the introduction of the verifiable step reward, GiGPO becomes similar to SRGPO with a small number of steps per group. Although ``VP+GiGPO(w. VPR)'' can achieve a similar success rate to ``VP+SRGPO'' during training, ``VP+SRGPO'' demonstrates much faster convergence and better stability.

Furthermore, to compare the navigation performance of agents after RFT, we evaluate the SR on the Embench-Nav test scenarios using the last-epoch models. As shown in \cref{tab:main_results}, the testing navigation SR of ``VP+SRGPO'' is 15.1 pp higher than that of ``VP+GiGPO(w. VPR)'', and 31.5 pp higher than that of ``VP+GRPO'', indicating that the agent trained with SRGPO possesses superior generalization performance. On the more complex VLN-CE environment, our method also outperforms GRPO and GiGPO, and compared to VLN-R1, which uses historical frames in its input, our method achieves significant improvements in SR, SPL, and NE, validating the scalability of SRGPO.

In addition, the experimental results also show that although SFT alone can effectively improve the navigation ability of the agent, it may negatively impact the performance of SRGPO. This phenomenon may be caused by the reduction in policy diversity after SFT, which in turn restricts the agent’s capacity for exploration during the RFT process.

We also compared different grouping strategies for SRGPO. As reported in \cref{tab:local-grouping}, two more localized alternatives (within-task and within-trajectory grouping) both underperform batch-level random grouping.
The reason is that overly easy or hard tasks yield little intra-group reward variance, whereas batch-level sampling preserves diversity by mixing steps from tasks of different difficulty.

% \begin{table}
%   \caption{Comparison of Different Methods (i.d.).}
%   \label{tab:method_comparison}
%   \centering
%   \begin{tabular}{@{}ccc@{}}
%     \toprule
%     \textbf{Method} & \textbf{Succ. rate} & \textbf{Avg. steps} \\
%     \midrule
%     Qwen2.5-VL-3B & 0.050 & 19.72 \\
%     +SFT & 0.367 & 16.63 \\
%     +GRPO & 0.400 & 16.32 \\
%     +GiGPO(vanilla) & 0.283 & 17.22 \\
%     +GiGPO(w. VPR) & 0.617 & 14.80 \\
%     \textbf{+SRGPO} & \textbf{0.717} & \textbf{13.77} \\
%     +SFT+SRGPO & 0.433 & 15.82 \\
%     \bottomrule
%   \end{tabular}
% \end{table}

% \subsection{Discussion of Hyperparameter for SRGPO}
% The step-level group size $N_S$ is an important hyperparameter in SRGPO. If adopting a small $N_S$, the estimation of step-level advantage may become less accurate and less robust, which in turn negatively affects the generalization ability of the VLN agent. We tried reducing $N_S$ from 16 to 8 and ran SRGPO for three times. The resulting agents attain a mean navigation success rate of 0.411 with a standard deviation of 0.103 on the test set, which is notably worse than the performance achieved with $N_S=16$ as reported in \cref{tab:main_results}. This phenomenon also explains why in \cref{fig:episode-success_rate-qwen3B} GiGPO with VPR can converge to a level comparable to SRGPO in the training scenarios, but its testing success rate still lags far behind SRGPO. Notably, increasing $N_S$ does not require any additional rollout computation cost in SRGPO, which demonstrates a more scalable advantage compared to GiGPO.

\subsection{Out-Of-Domain Generalization Capability}
To further evaluate the generalization ability of the model trained with SRGPO, we switched the training scenes of Embench-Nav to o.o.d. setting. During the training process, the navigation success rates of models trained with different algorithms on the test set change as illustrated in the right part of \cref{fig:embench_nav_curves}. From the figure, it can be seen that the testing curve of SRGPO is significantly higher than those of the other three algorithms, indicating that even when trained in o.o.d. scenes, the model trained by SRGPO demonstrates stronger navigation capabilities than the other models. This indicates that SRGPO enables the model to acquire more general navigation knowledge, confirming its superior generalization.

\section{Conclusion}
In this work, we propose SeeNav-Agent, a novel LVLM-based embodied navigation framework that includes a zero-shot dual-view visual prompt technique for the input side and an efficient RFT algorithm SRGPO for post-training. With the dual-view VP modules, the visual hallucinations can be significantly reduced and the spatial understanding capability of the VLN Agent can be improved. Through systematic analysis and ablation experiments, we identified the optimal VP recipe, which enables the best alignment and utilization of information from dual-view inputs. Based on the defined verifiable process reward for navigation that is independent of the state, the proposed SRGPO algorithm can randomly group steps for advantage estimation and thereby efficiently utilize the step reward signals. Compared with GRPO and GiGPO, SRGPO not only addresses their inherent limitations, but also demonstrates significant advantages in training stability, convergence efficiency, and generalization ability. On Embench-Nav, ``GPT4.1+VP'' achieves SR surpassing the previous SOTA close-source model by 20 pp, while ``VP+SRGPO'' exceeds the previous SOTA model by 5.6 pp. On VLN-CE, our method exceeds VLN-R1 by 8.9 pp in SR and outperforms it on SPL and NE, showing the superiority and scalability of SeeNav-Agent.

A limitation of our method is that the process reward proposed in this work is not applicable to complex navigation tasks that require traversing between floors. For such problems, designing more ingenious and general rewards is a worthwhile direction for future research. Although this work focuses on the navigation task, we may also consider extending the proposed SRGPO to other tasks where a state-independent process reward can be well-defined.

\bibliography{main}
\bibliographystyle{iclr2027_conference}

\appendix
% ==========================================================================
% Technical Appendix (from sec/X_suppl.tex of the ECCV source).
% ==========================================================================
\section{Experimental Details}
\label{sec:Experimental-Details}
\subsection{Parameters for EmbodiedBench-Navigation Environment}
The parameters for the EmbodiedBench-Navigation environment are set as follows, which are aligned with the original parameters used in the original EmbodiedBench paper \cite{yang2025embodiedbench}. These parameters are used in both training and testing environments.

\begin{itemize}
  \item \textbf{Visible Distance:} 10m --- The maximum visible distance for objects perceived by the agent.
  \item \textbf{Image Width:} 500px --- The pixel width of the single-view image.
  \item \textbf{Image Height:} 500px --- The pixel height of the single-view image.
  \item \textbf{Field of View:} 100$^\circ$ --- The field of view width visible from the agent's first-person perspective.
  \item \textbf{Move Magnitude:} 0.25m --- The distance the agent moves one step forward, backward, left, or right.
  \item \textbf{Rotate Degree:} 90$^\circ$ --- The number of degrees the agent turns its view left or right each time.
  \item \textbf{Tilt Degree:} 30$^\circ$ --- The number of degrees the agent turns its view up or down each time.
  \item \textbf{Min Distance:} 2.5m --- The minimum distance between the agent's starting position and the target.
  \item \textbf{Max Distance:} 3m --- The maximum distance between the agent's starting position and the target.
  \item \textbf{Success Threshold:} 1m --- The distance threshold for determining whether the agent has successfully reached the vicinity of the target.
  \item \textbf{Action Space:} eight low-level actions with IDs: Move 0-forward / 1-backward / 2-right / 3-left by $\Delta x$ meters; Rotate to the 4-right / 5-left by $\Delta \theta$ degrees; Tilt the camera 6-upward / 7-downward by $\Delta \phi$ degrees. Each task is defined by a unique initial robot pose, target object, and language instruction.
\end{itemize}

\subsection{Parameters for VLN-CE Environment}
The action space and parameters for the VLN-CE environment are set as follows. These parameters are used in both training and testing environments.
\begin{itemize}
  \item \textbf{Action Space:} Stop(0); Move forward(1); Turn left(2); Turn right(3).
  \item \textbf{Dataset Split:} 10,819 training samples and 1,839 val-unseen evaluation samples; scenes are based on real indoor environment scans.
  \item \textbf{Image Width:} 500px --- The pixel width of the single-view image.
  \item \textbf{Image Height:} 500px --- The pixel height of the single-view image.
  \item \textbf{Field of View:} 90$^\circ$ --- The field of view width visible from the agent's first-person perspective.
  \item \textbf{Move Magnitude:} 0.5m --- The distance the agent moves one step forward, backward, left, or right.
  \item \textbf{Rotate Degree:} 30$^\circ$ --- The number of degrees the agent turns its view left or right each time.
  \item \textbf{Success Threshold:} 3m --- The distance threshold for determining whether the agent has successfully reached the vicinity of the target.
\end{itemize}

\subsection{Training Protocol and Compute}
\label{sec:train-protocol}
For in-distribution (i.d.) training on Embench-Nav, we randomize the 60 base scenes by randomly placing movable objects, randomly setting the agent's initial position and orientation, and randomly selecting navigation goals. For out-of-distribution (o.o.d.) training, we select 60 new scenes from the AI2-THOR scene library that are not used in Embench-Nav, and apply the same randomization. Dual-view inputs are constructed by concatenating FV (left) and BEV (right) into a single image. All experiments were run on 8 H20 GPUs (96 GB VRAM each); a single RFT run took about 48/96 hours on Embench-Nav/VLN-CE.

\subsection{Example of Language Prompts on Embench-Nav}
The language prompts used for the SeeNav-Agent with visual prompts for Embench-Nav in this work is shown in \cref{tab:language_prompts}. In the prompt, \textless image\textgreater is a placeholder for the input image, \{user\_instruction\} is the navigation command given by human, like ``Navigate to the Bread in the room and be as close as possible to it'', and \{history\} represents the interaction history between the agent and the environment, which is a list of natural language. An example of the interaction history for Embench-Nav is shown in \cref{tab:history_example}. The maximum time window for the interaction history is set to 5 in this work, which means that the \{history\} part is limited to recording interaction information for the past five steps at most. When VP or certain modules of VP are not used in the experiments, we remove the prompts related to the corresponding modules from the original prompt.

\begin{longtable}{p{0.9\textwidth}}
\caption{\textbf{An Example of the Interaction History.}}
\label{tab:history_example}\\
\hline
 The action history:

 \vspace*{0.3cm}
 \textless action info\textgreater

 Step 0, action id 0, Move forward by 0.25.

 \textless env feedback\textgreater

 Last action MoveAhead executed successfully.

 \vspace*{0.3cm}
 \textless action info\textgreater

 Step 1, action id 0, Move forward by 0.25.

 \textless env feedback\textgreater

 Last action MoveAhead executed successfully.

 \vspace*{0.3cm}
 \textless action info\textgreater

 Step 2, action id 0, Move forward by 0.25.

 \textless env feedback\textgreater

 Last action MoveAhead executed successfully.

 \vspace*{0.3cm}
 \textless action info\textgreater

 Step 3, action id 2, Move rightward by 0.25.

 \textless env feedback\textgreater

 Last action MoveRight is invalid. Cube.668 is blocking Agent 0 from moving by (0.0000, 0.0000, 0.2500).

 \vspace*{0.3cm}
 \textless action info\textgreater

 Step 4, action id 3, Move leftward by 0.25.

 \textless env feedback\textgreater

 Last action MoveLeft is invalid. Cube.690 is blocking Agent 0 from moving by (0.0000, 0.0000, -0.2500).
 \\
\hline
\end{longtable}

% NOTE: The language prompt is very long; use longtable so it can span pages.
{\small
\begin{longtable}{L{0.9\textwidth}}
\caption{\textbf{Language Prompt Template for SeeNav-Agent with Visual Prompts.}}
\label{tab:language_prompts}\\
\toprule
\textcolor{blue}{\textless image\textgreater} \#\# You are a robot operating in a home. You can do various tasks and output a sequence of actions to accomplish a given task with images of your status.

Your input is a concatenation of your top-down view and the first-person view image, with the top-down image on the left and the first-person view on the right.

The colored circle in the top-down view represents your current position, with the YELLOW side indicating your LEFT and the PURPLE side indicating your RIGHT. The GREEN arrow shows your current camera orientation. In the first person view and the overhead view, the red bounding box in both views highlights the object you need to navigate to.
The candidate actions are shown in the overhead view and the first person view images, and each action is represented by a blue arrow, with the corresponding action ID at the end of the arrow.

\vspace*{0.3cm}
Now the human instruction is: \textcolor{blue}{\{user\_instruction\}}.

\vspace*{0.3cm}
\#\# The available action id (0 to 7) and action names are:
action id 0: Move forward by 0.25, \\
action id 1: Move backward by 0.25, \\
action id 2: Move rightward by 0.25, \\
action id 3: Move leftward by 0.25, \\
action id 4: Rotate to the right by 90 degrees, \\
action id 5: Rotate to the left by 90 degrees, \\
action id 6: Tilt the camera upward by 30 degrees, \\
action id 7: Tilt the camera downward by 30 degrees.

\vspace*{0.3cm}
Among these actions, actions 0-3 will only be annotated in the top-down view, while actions 4-7 will only be annotated in the first-person view.

\vspace*{0.3cm}
\textcolor{blue}{\{history\}}

\vspace*{0.3cm}
Now your task is to determine which arrow's action will help you reach the navigation target with a red bounding box (if you encounter any obstacles, prioritize bypassing the obstacles currently blocking your way), and choose the optimal action id.

\vspace*{0.3cm}
*** Strategy ***

To achieve the task, 1. Reason about the current visual state and your final goal, and 2. Reflect on the effect of previous actions. 3. Summarize how you learn from the Strategy. Aim for one action per step. At last, output the action id from the available actions to execute.\\

1. The action arrows on the images indicate the forward direction of the agent after performing the corresponding actions (actions 0-3), or the direction of the agent's view rotation (actions 4-7). You can use this information to help decide which action to take.\\

2. Strategy2: When determining the relative left-right position between the target and the agent, do not simply look at whether the target is on the left or right side of the top-down view. Instead, you need to take the agent’s orientation into account (for example, when the agent is facing downward in the image, objects that appear to be on the left side in the image are actually on the agent’s right side).\\

3. Strategy3: The red bounding box marks your navigation target. Please pay special attention to whether there is a corresponding red bounding box and a red nevigation arrow in your FIRST-PERSON VIEW. Avoid mistakenly judging an existing box as absent, or assuming a non-existent box is present.\\

4. Strategy4: There is a red navigation arrow in the top-down view and the first-person view that point from the agent to the target, you can use this arrow to assist the current navigation task. If the red navigation arrow is not visible in the first-person view, it means that the target is not visible in the first-person view.\\

5. Strategy5: When you choose an action, please pay attention to the relationship between the action arrows in the top-down view and the red navigation arrow. Your movement direction should help shorten the red navigation arrow and align the green arrow with the red arrow.\\

6. Strategy6: If an invalid action has occurred in the action history, please do not perform this action again unless you have already performed a rotation operation.\\

7. Strategy7: Based on the information in the images, determine which arrow's action will best help you reach the navigation target with a red bounding box (if you encounter any obstacles, prioritize bypassing the obstacles currently blocking your way), and choose the optimal action.\\

8. Strategy8: You should use rotation (action 4-5) or camera tilt (action 6-7) sparingly, only when you lose track of the target object and it's NOT IN YOUR VIEW. If so, plan nothing but ONE ROTATION OR TILT at a step until that object appears in your view. After the target object appears, start navigation and avoid using rotation until you lose sight of the target again.\\

\vspace*{0.3cm}
You are supposed to output in JSON.

The output json format should be \{`visual\_state\_description': str,
`reasoning\_and\_reflection': str, "language\_plan': str, `executable\_plan': List[\{`action\_id': int, `action\_name': str\}]\}

The fields in above JSON follows the purpose below:

1. visual\_state\_description is for description of current state of the first-person view and the top-down view image. \\

2. reasoning\_and\_reflection is for summarizing the history of interactions and any available environmental feedback. Additionally, provide reasoning as to why the last action or plan failed and did not finish the task, and providing reasoning process of why you choose the current action. \\

3. language\_plan is for describing the best action you choose from all the action arrows, which is started by the step number and the action name.\\

4. executable\_plan is the best action you choose that having an action ID and a name.\\

5. keep your plan efficient and concise.\\
\bottomrule
\end{longtable}
}

\end{document}